\documentclass[pdflatex,sn-mathphys-num]{sn-jnl}

\usepackage{graphicx}%
\usepackage{multirow}%
\usepackage{amsmath,amssymb,amsfonts}%
\usepackage{amsthm}%
\usepackage{mathrsfs}%
\usepackage[title]{appendix}%
\usepackage{xcolor}%
\usepackage{textcomp}%
\usepackage{manyfoot}%
\usepackage{booktabs}%
\usepackage{algorithm}%
\usepackage{algorithmicx}%
\usepackage{algpseudocode}%
\usepackage{listings}%
\usepackage{tabularx}
\usepackage{makecell} 

\theoremstyle{thmstyleone}%
\theoremstyle{thmstyletwo}%
\newtheorem{remark}{Remark}%

\theoremstyle{thmstylethree}%

\begin{document}

\title[Semantic Homogenization in Italian Popular Music: A Diachronic Analysis]{Semantic Homogenization in Italian Popular Music: A Diachronic Analysis}


\author*[1,2]{\fnm{Lorenzo} \sur{Canale}}\email{lorenzo.canale@rai.it}
\equalcont{These authors contributed equally to this work.}

\author*[1,3]{\fnm{Stefano} \sur{Scotta}}\email{stefano.scotta@rai.it}
\equalcont{These authors contributed equally to this work.}

\author*[1,4]{\fnm{Alberto} \sur{Messina}}\email{alberto.messina@rai.it}

\affil[1]{\orgdiv{Centro Ricerche, Innovazione Tecnologica e Sperimentazione}, 
\orgname{RAI}, 
\orgaddress{\street{Via Giovanni Carlo Cavalli 6}, \city{Turin}, \postcode{10138}, \country{Italy}}}

\affil[2]{ORCID: \href{https://orcid.org/0000-0002-7556-595X}{0000-0002-7556-595X}; Google Scholar: \href{https://scholar.google.com/citations?user=hsgYvg0AAAAJ}{hsgYvg0AAAAJ}}

\affil[3]{ORCID: \href{https://orcid.org/0000-0003-1078-2985}{0000-0003-1078-2985}; Google Scholar: \href{https://scholar.google.com/citations?user=mJRjh1sAAAAJ}{mJRjh1sAAAAJ}}

\affil[4]{ORCID: \href{https://orcid.org/0000-0002-8262-2449}{0000-0002-8262-2449}; Google Scholar: \href{https://scholar.google.com/citations?user=hmd7668AAAAJ}{hmd7668AAAAJ}}

\abstract{In recent years, studies have revealed a decline in semantic variety across popular music lyrics, particularly in English-language songs on streaming platforms like Spotify. This research examines whether a similar trend can be observed in a different linguistic and cultural context: the lyrics of all finalist songs from the 75 editions of the Sanremo Music Festival, Italy’s most renowned music competition. What sets this work apart is the development of a flexible and efficient methodology for tracking changes in semantic similarity over time, which can be applied to different datasets to study similar phenomena. Drawing on a combination of full-text, segment-based, topic-based, and word-level analyses, the approach leverages both embedding techniques and large language models. When applied to the Sanremo corpus, this framework reveals a gradual move toward increasing semantic uniformity, echoing the global patterns identified in previous studies. These findings underscore the value of natural language processing tools in uncovering long-term shifts in musical language and cultural expression.}




\maketitle

\noindent\textit{This preprint has not undergone peer review (when applicable) or any post-submission improvements or corrections. The Version of Record of this article is published in Journal of Computational Social Science, and is available online at \href{https://doi.org/10.1007/s42001-026-00468-1}{https://doi.org/10.1007/s42001-026-00468-1}.}

\vspace{1em}

\section{Introduction}

\begin{figure*}[!htb]
    \centering
    \includegraphics[width=0.78\linewidth]{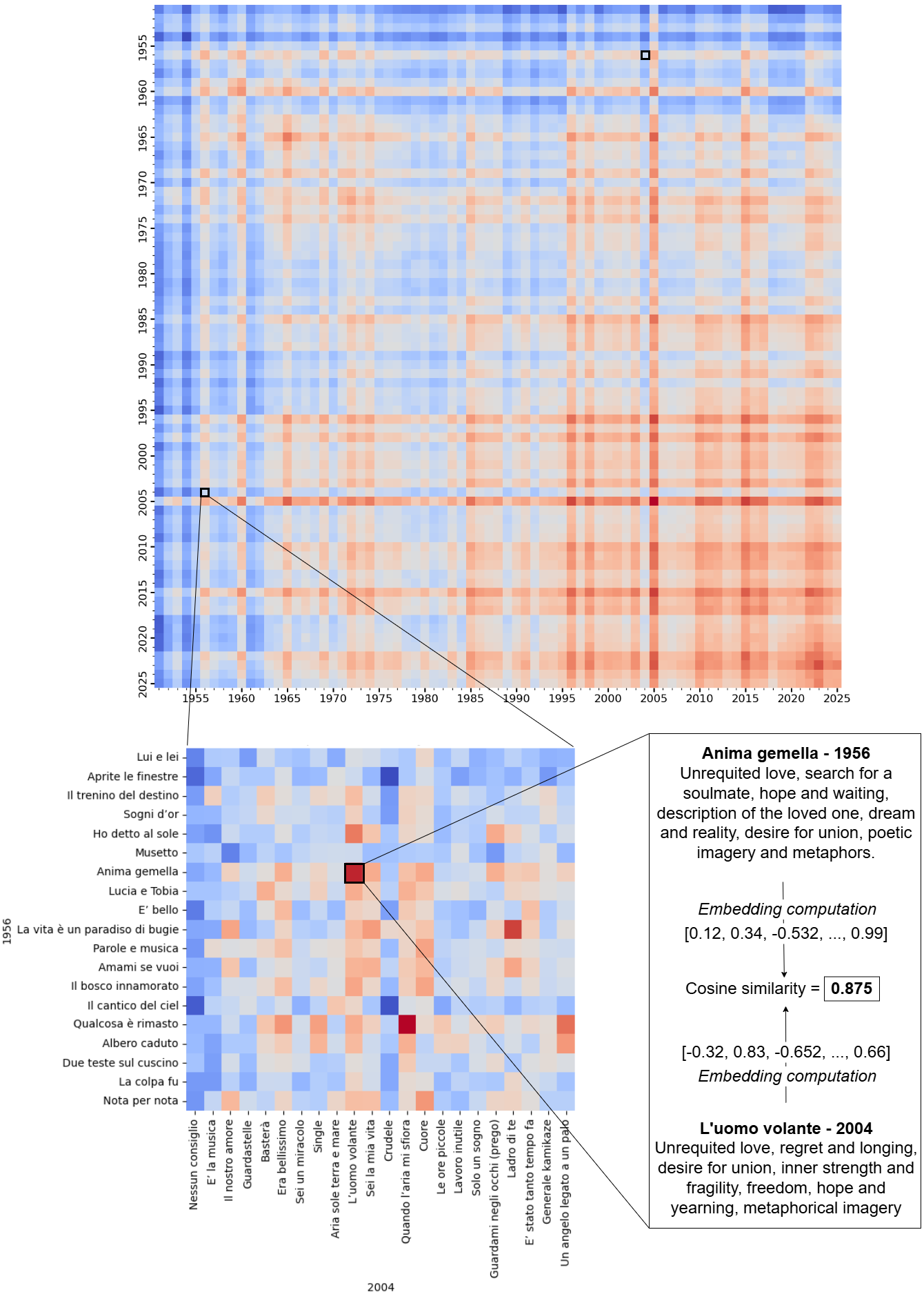}
    \caption{The image provides an overview of the methodology presented in the paper, illustrating how semantic similarity is analyzed across different years of the Sanremo Music Festival. The top part of the image displays a macro-matrix representing the pairwise similarity between all years in the dataset, calculated using a topic-based method. Each cell represents a pair of years, with color intensity indicating average topic similarity (extracted via Gemini from lyrics). The bottom left quadrant zooms in on a specific pair of years, 1956 and 2004, as an example of the detailed pairwise year analysis conducted. It shows a sub-matrix where each cell represents the topic similarity between every song from 1956 (rows) and 2004 (columns). To the right, an example compares a 1956 song ("Anima gemella") with one from 2004 ("L'uomo volante"), illustrating individual-level analysis. For these two songs, the English translations of the topics are shown. Embedding vectors of the Italian topic descriptions are computed, and their cosine similarity is calculated. A similar procedure, adapted for each method, was applied to the other four semantic similarity analyses; for example, Full-Text Similarity uses complete lyrics for embedding and similarity calculations.}
    \label{fig:figura1}
\end{figure*}

Understanding textual meaning has long been a central concern in both literary studies and computational linguistics. In literary theory, semantics is the study of how words and structures convey meaning, encompassing aspects such as denotation, connotation, and intertextuality \cite{cd470fd1e7f149fc98e0b0e535f7573c,Panagiotidou+2012+47+65}. Computational approaches to semantics aim to model these nuances through mathematical representations, with word and sentence embeddings emerging as a powerful technique to capture semantic similarities between texts \cite{harispe2022semantic,rapaport1994syntactic}.

Semantic embeddings map textual units (e.g., words, sentences, or entire documents) into high-dimensional vector spaces, where distances between vectors correspond to semantic similarity \cite{li2020sentence}. Models such as Word2Vec \cite{word2vec}, GloVe \cite{glove}, and transformer-based embeddings \cite{kalyan2021ammus} have demonstrated remarkable capabilities in capturing linguistic patterns, including synonymy, topical coherence, and stylistic variation.

In this study, we present two main contributions:
\begin{itemize}
\item A methodology that enables the analysis of semantic similarity within a musical collection by selecting specific variables for comparison, such as time. By combining different similarity measures, such as full-text, portion-based, topic-based, and word-based similarity—using embedding models and large language models (LLMs), this approach provides a more comprehensive analysis.

\item An in-depth analysis of the Sanremo Festival lyrics over time: we apply the methodology to analyze lyrical diversity at the Sanremo Festival, a major Italian music competition. Our findings indicate a trend of increasing semantic homogeneity in lyrics over time, suggesting a shift in artistic expression that may reflect broader cultural and industry-wide influences.
\end{itemize}

\section{Motivation}
Apart from its inherent entertainment nature, the Sanremo Festival \cite{Ghezzi2007}, being one of Italy's most prestigious and longest-running music competitions,  has long served as a significant cultural event in Italy, reflecting and influencing the country's social, political, and cultural landscapes and capturing the nation's attention during the week of the show. This is proved by the huge audience rates the Festival collects over the years, making it the most important annual television event. This prominence as a social event has fostered several studies, which highlight how the festival has evolved from reinforcing traditional values to addressing contemporary societal issues \cite{agostini2007, pierce2023, ardizzoni2020} as well as acting as a vehicle for political messages and populism-orientated discourse on pressing national and global topics \cite{caiani2023, way2016} and a platform to discuss important themes related to psychological health issues in modern society \cite{muller2025sanremo}.

It should be therefore evident that this study, conducted on the evolution and similarity of lyrics for the entire series of the Sanremo Festival, adds significant value to the current state of the art by providing a case study of long-term, genre-specific lyrical trends and similarities within a highly influential cultural event. Furthermore, while many studies explore the evolution of lyrics in popular music globally, a study focusing on the Sanremo Festival provides a specific cultural lens. By analyzing the lyrics of songs performed over decades at Sanremo, it is possible to track shifts in language use, thematic focus, and emotional expression within the context of Italian culture, offering a deeper understanding of the Italian musical landscape.

\section{Related work}

Recent research has highlighted a marked simplification in the musical structure of popular songs over time. For instance, Di Marco et al.~\cite{dimarco2025decodingmusicalevolutionnetwork} analyzed over 20,000 compositions using tools from network science, revealing a significant reduction in musical complexity across diverse genres. Similarly, Hamilton and Pearce~\cite{Hamilton} examined U.S. chart-topping melodies from 1950 to 2023 and identified a decreasing trend in melodic variety and unpredictability. While these studies focus on the structural and harmonic dimensions of music, our work diverges by addressing the evolution of lyrics.

Previous approaches to analyzing musical texts have applied various methods, including word2vec and bag-of-words. For instance, \cite{9077706} focused on rap lyrics, using word embeddings to capture semantic meaning. On the other hand, \cite{8614228} introduced graph-based representation to analyze and interpret popular music lyrics.

In this paper, a key factor driving the adoption of transformer embeddings and semantic similarity measures is their alignment with human perception. In particular, \cite{kim2024computationalanalysislyricsimilarity} recently found a significant correlation between semantic similarity scores and perceptual similarity judgments, suggesting that embedding-based approaches can effectively approximate human cognitive evaluations of textual similarity; while they utilized the \verb|all-MiniLM-L6-v2|\footnote{\url{https://huggingface.co/sentence-transformers/all-MiniLM-L6-v2}} model, in our analysis we use more recent embedding models (see Section \ref{dataemb}) because of their superior performance in benchmark evaluations\footnote{\url{https://huggingface.co/spaces/mteb/leaderboard}} (see \cite{MTEB}).

In \cite{poetrygerman} the authors employed embeddings also to trace trends in textual evolution, highlighting the increasing significance of semantic embeddings in the humanities. Their work supports some hypotheses from literary studies while offering fresh perspectives on the development of early modernism in German poetry.

Furthermore, in \cite{Parada-Cabaleiro2024} it was discovered that lyrics have become simpler over time across several dimensions. While the authors analyzed lyrical trends using handcrafted descriptors—such as lexical diversity, structural patterns, and rhyme characteristics—our approach differs by utilizing semantic embeddings. At the conclusion of their study, the authors emphasize the broader implications of lyrical analysis, noting that song lyrics remain an underexplored but valuable resource for understanding cultural artifacts and societal shifts. Our research aligns with this perspective, demonstrating how semantic similarity analysis can reveal evolving lyrical patterns, thus contributing to the broader discourse on cultural and artistic transformations in music. Additionally, our findings, derived from a prominent Italian music festival, reaffirm the trends observed in their more general dataset.

\section{Overview}
This work is structured as follows. Section \ref{dataemb} introduces the dataset we collected and the embedding models used throughout the experiments, along with the corresponding notation. Section \ref{analysis through texts} presents the techniques employed to analyze the evolution of semantic similarity over the years in the lyrics of songs performed at the Sanremo Festival finals. Each subsection focuses on a specific type of analysis.  See Figure \ref{fig:figura1} for an example that shows in general how the analysis is performed. Finally, in Section \ref{resultscom}, we discuss the results of the experiments described in Section \ref{analysis through texts}. Section \ref{future} concludes with a brief discussion of potential future work in this area.

\section{Dataset and Embeddings Used}\label{dataemb}
For each year $Y \in \mathcal{Y} = \{1951, \dots, 2025\}$ in which the Sanremo festival took place, we collected the lyrics of the $n_Y$ finalist songs, denoted as $l^Y_1, \dots, l^Y_{n_Y}$.

To analyze the similarities between songs, we considered a set $E$ of different textual embedding models. Specifically, we considered $E = \{i, o, c\}$, where:
\begin{itemize}
\item $i$ represents the sentence transformer model \verb|intfloat/multilingual-e5-large|\footnote{\url{https://huggingface.co/intfloat/multilingual-e5-large}}, see \cite{e5} for details;
\item $o$ corresponds to OpenAI’s embedding model \verb|text-embedding-3-large|\footnote{\url{https://openai.com/index/new-embedding-models-and-api-updates/}},
\item $c$ denotes \verb|colbert-xm|\footnote{\url{https://huggingface.co/antoinelouis/colbert-xm}}, see \cite{colbert, colbert2} for details.
\end{itemize}

Note that hereinafter we will generally talk about similarity between embeddings. When the embeddings considered are $i$ and $o$ we call similarity the standard cosine similarity. When the embedding considered is $c$, since the ``MaxSim'' similarity $S$, commonly used for $c$, is not symmetric (see \cite{colbert, colbert2}), we call similarity between two $c$-embeddings $A$ and $B$, the quantity $\frac{1}{2}[S(A,B) + S(B,A)]$.

\section{Cross-Year Analysis of Semantic Similarity}\label{analysis through texts}
In this section, we present different approaches for analyzing the similarities between songs across the years. The goal is to compute, for each pair of years $(Y_a, Y_b) \in \mathcal{Y}\times \mathcal{Y}$, a measure that quantifies the average semantic similarity between the lyrics $l_{Y_a}^1, \dots, l_{Y_a}^{n_{Y_a}}$ and $l_{Y_b}^1, \dots, l_{Y_b}^{n_{Y_b}}$. This allows us to examine the evolution of lyrical similarity over time.

\subsection{Full-Text Similarity}\label{full_text}
\begin{figure}[h]
\centering
\includegraphics[width=0.75\columnwidth]{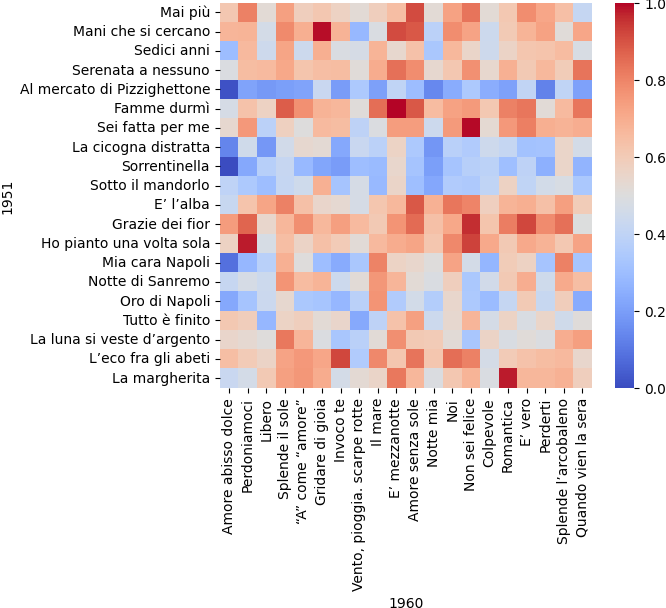}
\caption{Example of the similarity matrix $F^i_{1951,1960}$, where values are normalized such that the minimum is 0 and the maximum is 1. Rows and columns correspond to the titles of the finalist songs from 1951 and 1960, respectively. This grid shows the similarity in the full text of each song pairing, with warmer colors indicating more alike lyrics and cooler colors indicating less similarity. The normalization allows for easy comparison of relative similarities.}
\label{fig:sim_couple_years}
\end{figure}

We first analyze song similarities using embeddings computed from the full lyrics. Given an embedding model $\alpha \in E$, we construct, for each pair of years $(Y_a, Y_b) \in \mathcal{Y}$, an $n_{Y_a} \times n_{Y_b}$ similarity matrix $F^\alpha_{Y_a, Y_b}$, where each entry represents the cosine similarity between the embeddings (obtained with $\alpha$) of the songs  from $Y_a$ and $Y_b$. An example is shown in Figure \ref{fig:sim_couple_years}.

\begin{figure}[!ht]
    \centering
    \includegraphics[width=0.75\linewidth]{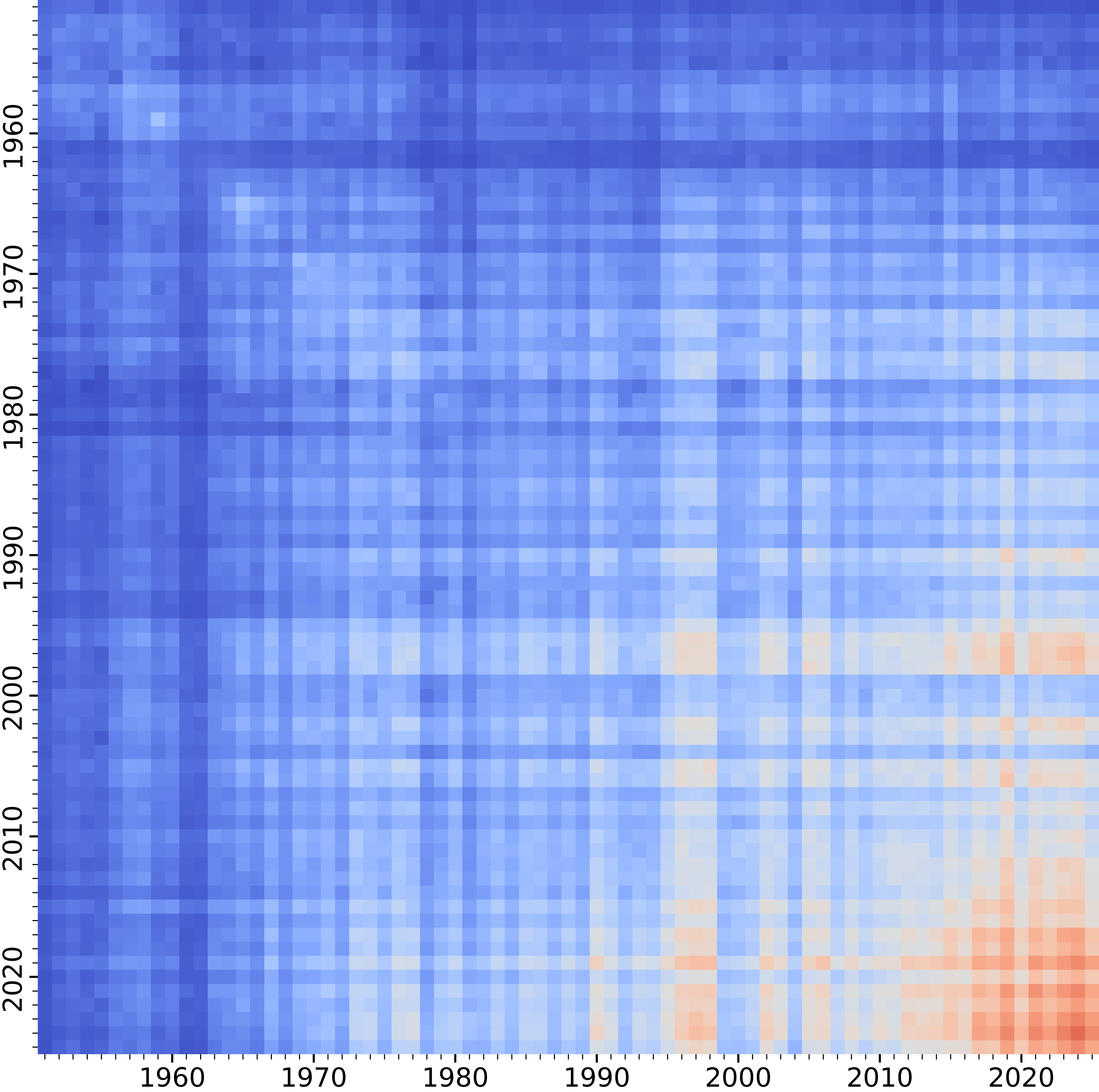}
    \caption{Matrix $\overline{F}^{i}_{0.75}$, as defined in Section \ref{full_text}, shows the percentage of song pairs between any two years where the full lyrics exceed the similarity of 75\% of all other song pairs. Warmer colors indicate a larger proportion of unusually high lyrical similarity between those years.}
    \label{fig:fulltext_intfoloate5_perc_nonorm}
\end{figure}

Next, we compute similarity measures for all possible year pairs $(Y_a, Y_b) \in \mathcal{Y} \times \mathcal{Y}$, avoiding the diagonal when $Y_a = Y_b$. Specifically, for each embedding model $\alpha \in E$, we derive the following matrices:

\begin{itemize}
\item $\overline{F}^{\alpha}_{mean}$ which is built evaluating for each $(Y_a, Y_b) \in \mathcal{Y}\times \mathcal{Y}$, the mean of all values in $F^{\alpha}_{Y_a,Y_b}$ (excluding the diagonal if $Y_a = Y_b$) obtaining a first $|\mathcal{Y}|\times|\mathcal{Y}|$ matrix. $\overline{F}^{\alpha}_{mean}$ is obtained normalizing the latter matrix in such a way that that the minimum value is $0$ and the highest is $1$.
\item $\overline{F}^{\alpha}_{med}$ which is analogous to $\overline{F}^{\alpha}_{mean}$ but considering the median (instead of the mean) similarity score, then normalized as $\overline{F}^{\alpha}_{mean}$.
\item $\overline{F}^{\alpha}_{q}$ whose element corresponding to $(Y_a, Y_b) \in \mathcal{Y}\times \mathcal{Y}$ is the proportion of elements of $F^{\alpha}_{Y_a,Y_b}$ that exceeds the $q$-quantile threshold, computed across all song pairs in the dataset. Selecting a high value of $q$, it represents the percentage of song pairs for each couple of years which are more similar than the majority of the other possible pairs. 

\end{itemize}
See Figure \ref{fig:fulltext_intfoloate5_perc_nonorm} for an example.

\subsection{Portion-Based Similarity}\label{portions}
A potential limitation of full-text analysis is that two songs might be highly similar even if their overall similarity score is not particularly high. Instead of comparing entire lyrics, we introduce a measure that assigns greater importance to similar textual segments.

 For each song $l_Y^i$, where $Y \in \mathcal{Y}$ and $i \in \{1, \dots, n_Y\}$, we segment it into portions $p(l_Y^i) = {p_1, \dots, p_{m_Y^i}}$, where $m_Y^i \geq 1$, ensuring that $l_Y^i$ is reconstructed from the ordered union of this portions. The portions are extracted using an LLM (namely \verb|gemini-1.5-flash|\footnote{\url{https://ai.google.dev/gemini-api/docs/models}}) prompted to segment lyrics into portions made up of more than one sentence. Hence, to evaluate the similarity between two songs  $l_{Y_a}^i$ and $l_{Y_b}^j$, for some $Y_a, Y_b \in \mathcal{Y}$ and $i \in \{1, \dots, n_{Y_a}\}; j \in \{1, \dots, n_{Y_b}\}$, we consider their portions $p(l_{Y_a}^i)$ and $p(l_{Y_b}^j)$ and compute the similarity between the $\alpha$-embedding of all the couples $(p_i, p_j)$, for $\alpha \in E$, $p_i \in p(l_{Y_a}^i)$ and $p_j \in p(l_{Y_b}^j)$, obtaining a $m_{Y_{a}}^i\times m_{Y_{b}}^j$ matrix $\Pi^{\alpha}(l_{Y_a}^i, l_{Y_b}^j)$. The element $(u,v)$ of the matrix $\Pi^{\alpha}(l_{Y_a}^i, l_{Y_b}^j)$ is the similarity between the $\alpha$-embedding of the $u$-th portion of $l_{Y_a}^i$ and the one of the $v$-th portion of $l_{Y_b}^j$. A concrete example of similar portions of text could be found in Table \ref{tab:lyric_similarity_segments}.

 Then, to visualize similarities between songs for a couple of years $(Y_a, Y_b) \in \mathcal{Y}\times\mathcal{Y}$ based on their parts, we define the $n_{Y_a}\times n_{Y_b}$ matrix $P^{\alpha, k}_{Y_a,Y_b}$, for some arbitrary small (see Remark \ref{k_value}) positive integer $k$, whose element $(i,j)$ is the sum of the $k$ highest values of the matrix $\Pi^{\alpha}(l_{Y_a}^i, l_{Y_b}^j)$. In the matrices $P^{\alpha, k}_{Y_a,Y_b}$ with $Y_a = Y_b$ we avoid the diagonal values. The matrix $P^{\alpha, k}_{Y_a,Y_b}$ is the analogous, but reasoning on songs'portions, of the matrix $F^{\alpha}_{Y_a,Y_b}$ (full-text similarity).

\begin{figure}[!t]
    \centering
    \includegraphics[width=0.75\linewidth]{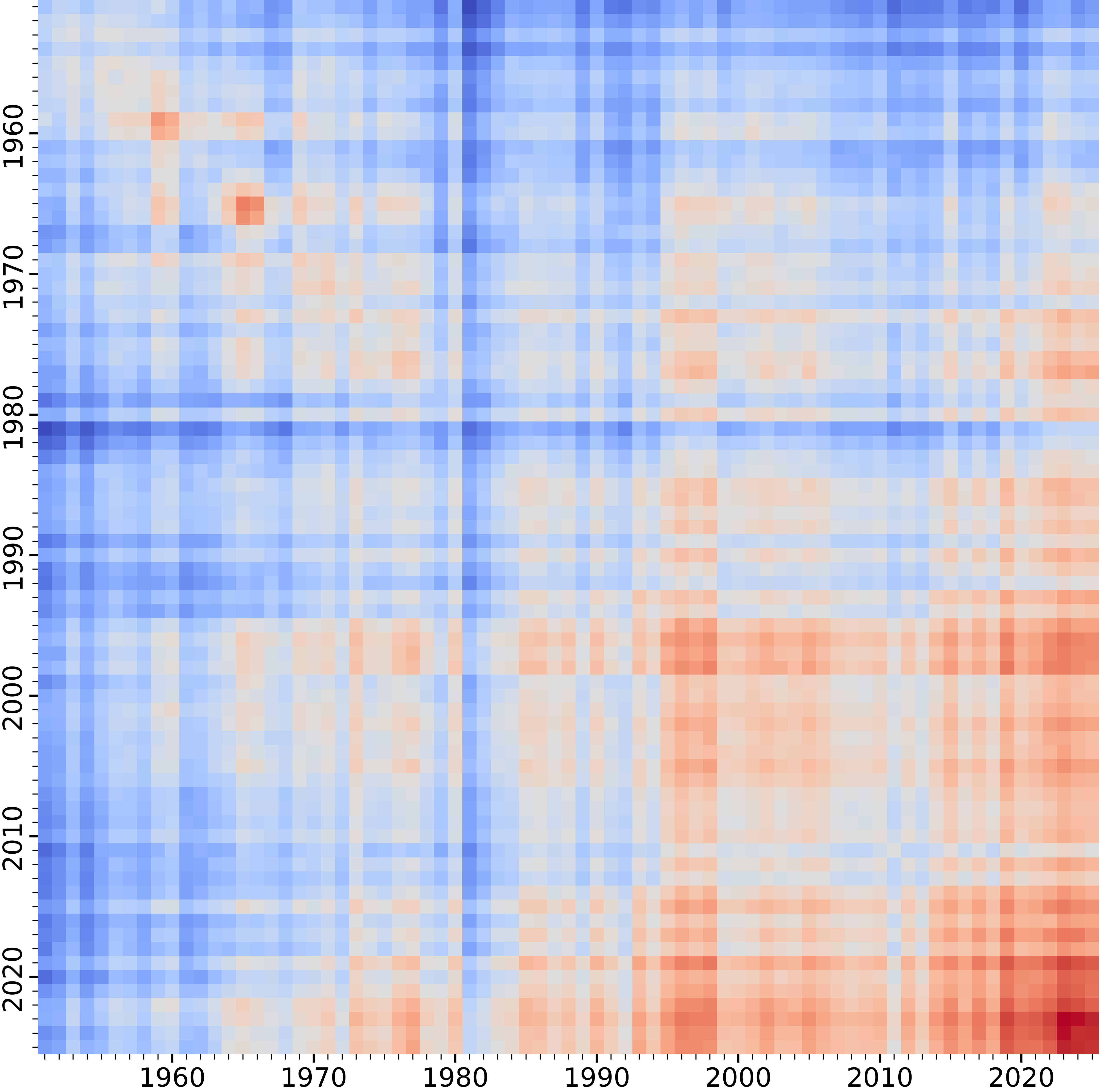}
\caption{Matrix $\overline{P}^{o, 3}_{mean}$, as defined in Section \ref{portions}, presents the average similarity based on the top 3 most similar sections of lyrics between songs from any two years. Warmer colors suggest a greater homogeneity in the lyrical content, specifically within the most comparable segments, between the songs of those two years.}

    \label{fig:portion_openai_mean_norm}
\end{figure}

 As we did in Section \ref{full_text}, in order to have matrices that allow us to visualize some similarity measures across years, we define the following (normalized) $|\mathcal{Y}|\times|\mathcal{Y}|$ matrices $\overline{P}^{\alpha, k}_{mean}$, $\overline{P}^{\alpha, k}_{med}$, $\overline{P}^{\alpha, k}_{q}$, which are defined for any $\alpha \in E$, exactly as, respectively, $\overline{F}^{\alpha}_{mean}$, $\overline{F}^{\alpha}_{med}$, $\overline{F}^{\alpha}_{q}$ but using the matrices $P^{\alpha, k}_{Y_a,Y_b}$ instead of $F^{\alpha}_{Y_a,Y_b}$, for $(Y_a, Y_b) \in \mathcal{Y} \times \mathcal{Y}$. See Figure \ref{fig:portion_openai_mean_norm} for an example.

 \begin{remark}\label{k_value}
    $k$ must be smaller than the minimum of $\{m_{Y_{a}}^im_{Y_{b}}^j: i \in \{1,\dots, n_{Y_a}\}, j \in\{1,\dots, n_{Y_b}\}\}$.
\end{remark}

\subsection{Topic-Based Similarity}\label{topicbased}
\begin{figure}[!b]
    \centering
    \includegraphics[width=0.75\linewidth]{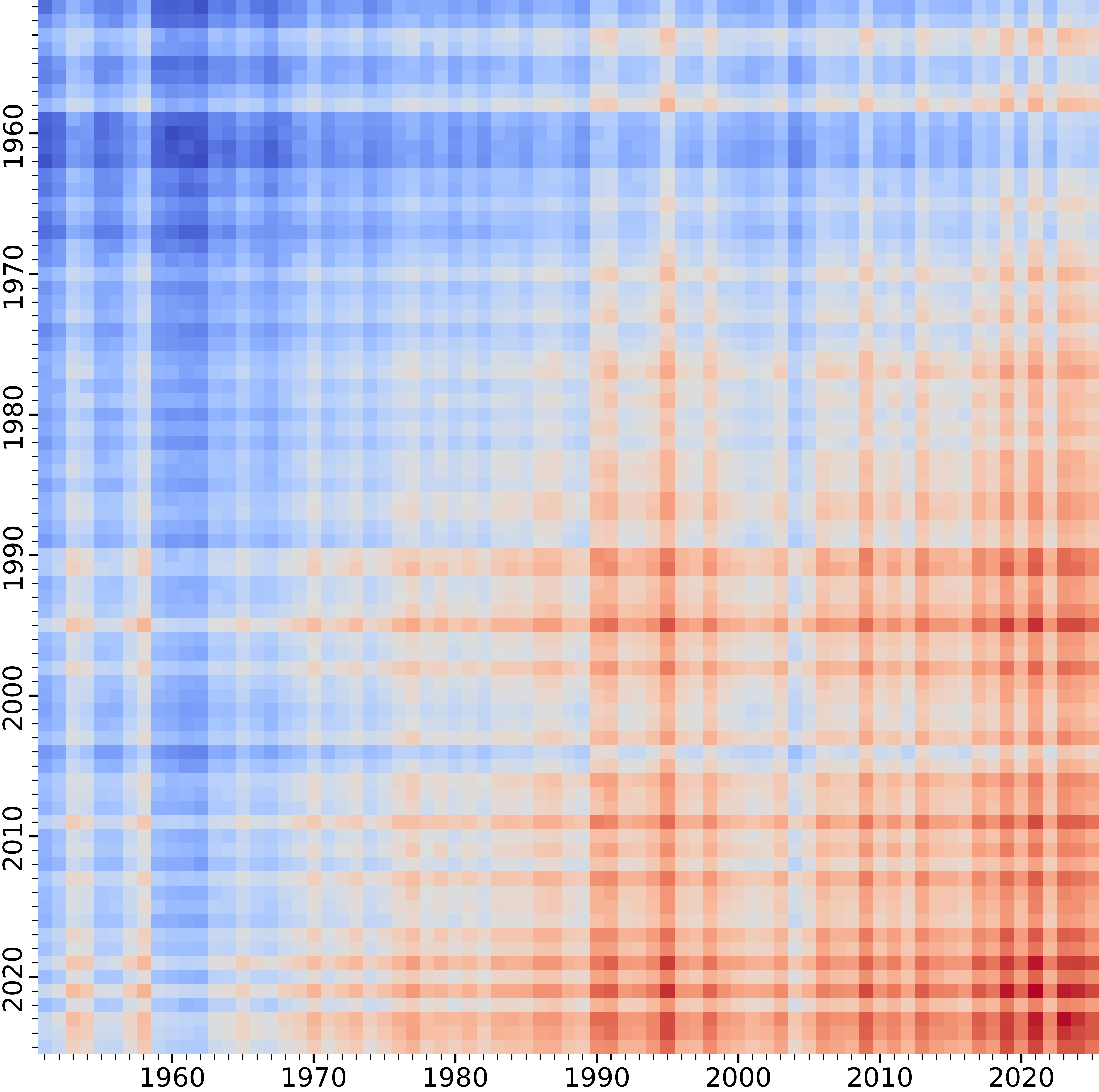}
    \caption{Matrix $\overline{T}^{c}_{med}$, as defined in Section \ref{topicbased}, illustrates the median similarity of the main themes discussed in songs between any two years. Warmer colors indicate a higher  thematic similarity.}
    \label{fig:topic_colbert_median_norm}
\end{figure}

Another approach to evaluating semantic similarity is through the analysis of lyrics topics.  To achieve this, in a first step we identify the primary topics $t^Y_i$ for each song $l^Y_i$, where $Y \in \mathcal{Y}$ and $i \in \{1, \dots, n_Y\}$. The identification process is done using a LLM (namely \verb|gemini-1.5-flash|) prompted to, given the lyrics of a song, return a list of the main topics treated in the text.
Once topics are extracted for each song, we compute the similarity between the topics $t^{Y_a}_i$ and $t^{Y_b}_j$ of two generic songs $l^{Y_a}_i$ and $l^{Y_b}_j$, for $Y_a, Y_b \in \mathcal{Y}$ and $i \in \{1,\dots, n_{Y_a}\}, j \in  \{1,\dots, n_{Y_b}\}$. This is done by converting the lists of topics into strings, which are then embedded using $\alpha \in E$. The similarity between two songs based on the topics is thus the similarity between the $\alpha$-embedding of these strings.
Therefore it is straightforward to build, for each $(Y_a,Y_b) \in \mathcal{Y}\times\mathcal{Y}$, the $n_{Y_a}\times n_{Y_b}$ matrix $T^\alpha_{Y_a, Y_b}$ containing the topic similarities between all the song presented at the Sanremo festival in $Y_a$ and $Y_b$ (avoiding the diagonal if $Y_a = Y_b$). $T^\alpha_{Y_a, Y_b}$ is analogous to $F^\alpha_{Y_a, Y_b}$ and $P^{\alpha, k}_{Y_a, Y_b}$ (defined in Sections \ref{full_text} and \ref{portions}), but topics based. Once we compute all the matrices $T^\alpha_{Y_a, Y_b}$, in order to analyze the evolutions of topics similarity across years, we compute the (normalized) $|\mathcal{Y}|\times |\mathcal{Y}|$ matrices $\overline{T}^{\alpha}_{mean}$, $\overline{T}^{\alpha}_{med}$, $\overline{T}^{\alpha}_{q}$ exactly as done in Section \ref{full_text} for, respectively, $\overline{F}^{\alpha}_{mean}$, $\overline{F}^{\alpha}_{med}$, $\overline{F}^{\alpha}_{q}$ but using the matrices $T^\alpha_{Y_a, Y_b}$ instead of $F^{\alpha}_{Y_a,Y_b}$, for $(Y_a, Y_b) \in \mathcal{Y} \times \mathcal{Y}$. See Figure \ref{fig:topic_colbert_median_norm} for an example.

\subsection{Word-Based Similarity}\label{wordsbased}
In this section, we adopt a different analytical approach, shifting from embedding-based models to a purely statistical evaluation of word occurrences. By focusing on words co-occurrence patterns across lyrics and years, it is possible to assess whether the results gained through embedding-based techniques align with the statistical ones.

For each $l^i_Y$, for $Y \in \mathcal{Y}$ and $i \in \{1,\dots, n_Y\}$, we considered both the lyric as-is and a normalized one $\tilde l^i_Y$ obtained converting all verbs to their infinitive form, transforming all text to lowercase, and reducing all nouns and adjectives to their singular form. The normalized text $\tilde l^i_Y$ is obtained prompting a LLM (namely \verb|gemini-1.5-flash|) to transform the text $l^i_Y$ as explained above.

\begin{remark}
    Although in principle the normalization of the text to pass from a lyric $l^i_Y$ to its normalized version $\tilde l^i_Y$  could be done even relying on simpler methods (like the spaCy\footnote{\url{https://spacy.io/}} python library \cite{spacy}) rather than LLM, we observed that even when using Italian-specific models, the resulting forms were often suboptimal—especially for verbs, which were lemmatized into unexpected or non-infinitive forms. For this reason, we opted for an LLM-based approach, which, in our experiments, produced more consistent and intuitive results for the data at hand. However, this was not a formal evaluation, but rather a practical choice based on manual inspection.
\end{remark}


To focus only on significant words, we filter each lyric (both original and normalized) to retain only verbs, nouns, and adjectives. This filtering is based on automatic part-of-speech (PoS) tagging performed using the \texttt{spaCy} Python library with the \texttt{it\_core\_news\_lg} model, which is specifically trained for Italian. The extraction is carried out in two ways: preserving repeated words or not, using the transformations $\rho$ and $\tilde\rho$ on a generic text $T$:
\begin{itemize}
    \item $\rho(T) =$ list of all significant words in $T$ with repetition;
    \item $\tilde\rho(T) =$ list of all significant words in $T$ without repetition.
\end{itemize}

So, for example, if the initial text is $T=$ ``my Home is a very sweet home'', its cleaned version $\tilde T$ would be ``my home be a very sweet home''. Then $\rho(T) = $ [Home, is, sweet, home] $ =\tilde\rho(T) $, $\rho(\tilde T) = $ [home, be, sweet, home], $\tilde\rho(\tilde T) = $ [home, be, sweet]. 

In this case, the ``similarity'' between two lyrics $l^i_{Y_a}$ and $l^j_{Y_b}$, for $Y_a, Y_b \in \mathcal{Y}$ and $i \in \{1,\dots, n_{Y_a}\}, j \in \{1,\dots, n_{Y_b}\}$, is given by $\rho(l^i_{Y_a}) \cap \rho(l^j_{Y_b})$ (resp. $\rho(\tilde l^i_{Y_a}) \cap \rho(\tilde l^j_{Y_b})$ on the normalized text), if we want to give more weight to the repeated words, and by $\tilde\rho(l^i_{Y_a}) \cap \tilde\rho(l^j_{Y_b})$ (resp. $\tilde\rho(\tilde l^i_{Y_a}) \cap \tilde\rho(\tilde l^j_{Y_b})$) otherwise.

Now, that we defined the similarity for this type of analysis, we build for each pair of years $(Y_a, Y_b) \in \mathcal{Y}\times\mathcal{Y}$ four different matrices $W_{Y_a, Y_b}^{\iota}$, $\iota \in \{1,2,3,4\}$, containing the similarities between the songs in $Y_a$ and $Y_b$ defined as above, considering the four combination of as-is or normalized text and intersection with or without repetition. More precisely, the similarity between $l^i_{Y_a}$ and $l^j_{Y_b}$ will be represented in $W_{Y_a, Y_b}^{1}$, $W_{Y_a, Y_b}^{2}$, $W_{Y_a, Y_b}^{3}$, $W_{Y_a, Y_b}^{4}$ as, respectively, $\rho(l^i_{Y_a}) \cap \rho(l^j_{Y_b})$, $\tilde\rho(l^i_{Y_a}) \cap \tilde\rho(l^j_{Y_b})$, $\rho(\tilde l^i_{Y_a}) \cap \rho(\tilde l^j_{Y_b})$, $\tilde\rho(\tilde l^i_{Y_a}) \cap \tilde\rho(\tilde l^j_{Y_b})$. This is summarized in Table \ref{tab:rho_transformations}.

\begin{table}[h]
    \caption{Notation for the $W_{Y_a, Y_b}^{\iota}$, $\iota \in \{1,2,3,4\}$ and $(Y_a, Y_b) \in \mathcal{Y}\times\mathcal{Y}$, obtained through the intersections of $\rho$ and $\tilde \rho$  transformations, applied to lyrics in their original (``as-is'') and normalized forms.}
    \label{tab:rho_transformations}
    \begin{tabular}{c|cc}
        & $\rho$ & $\tilde{\rho}$ \\
        \hline
        \rule{0pt}{12pt} as-is & $W^1_{Y_a, Y_b}$ & $W^2_{Y_a, Y_b}$ \\
       \rule{0pt}{12pt} normalized & $W^3_{Y_a, Y_b}$ & $W^4_{Y_a, Y_b}$ \\
    \end{tabular}
\end{table}

Then, to analyze the word-based similarities, defined above, across the years, we proceed as in the previous sections, see Section \ref{full_text} for details, to calculate, for each $\iota \in \{1,2,3,4\}$ the (normalized) $|\mathcal{Y}|\times |\mathcal{Y}|$ matrices $\overline W^{\iota}_{mean}$, $\overline W^{\iota}_{med}$, $\overline W^{\iota}_{q}$ from the matrices $W_{Y_a, Y_b}^{\iota}$ for all $(Y_a, Y_b) \in \mathcal{Y}\times\mathcal{Y}$.

\begin{figure}[!htp]
    \centering
    \includegraphics[width=0.75\linewidth]{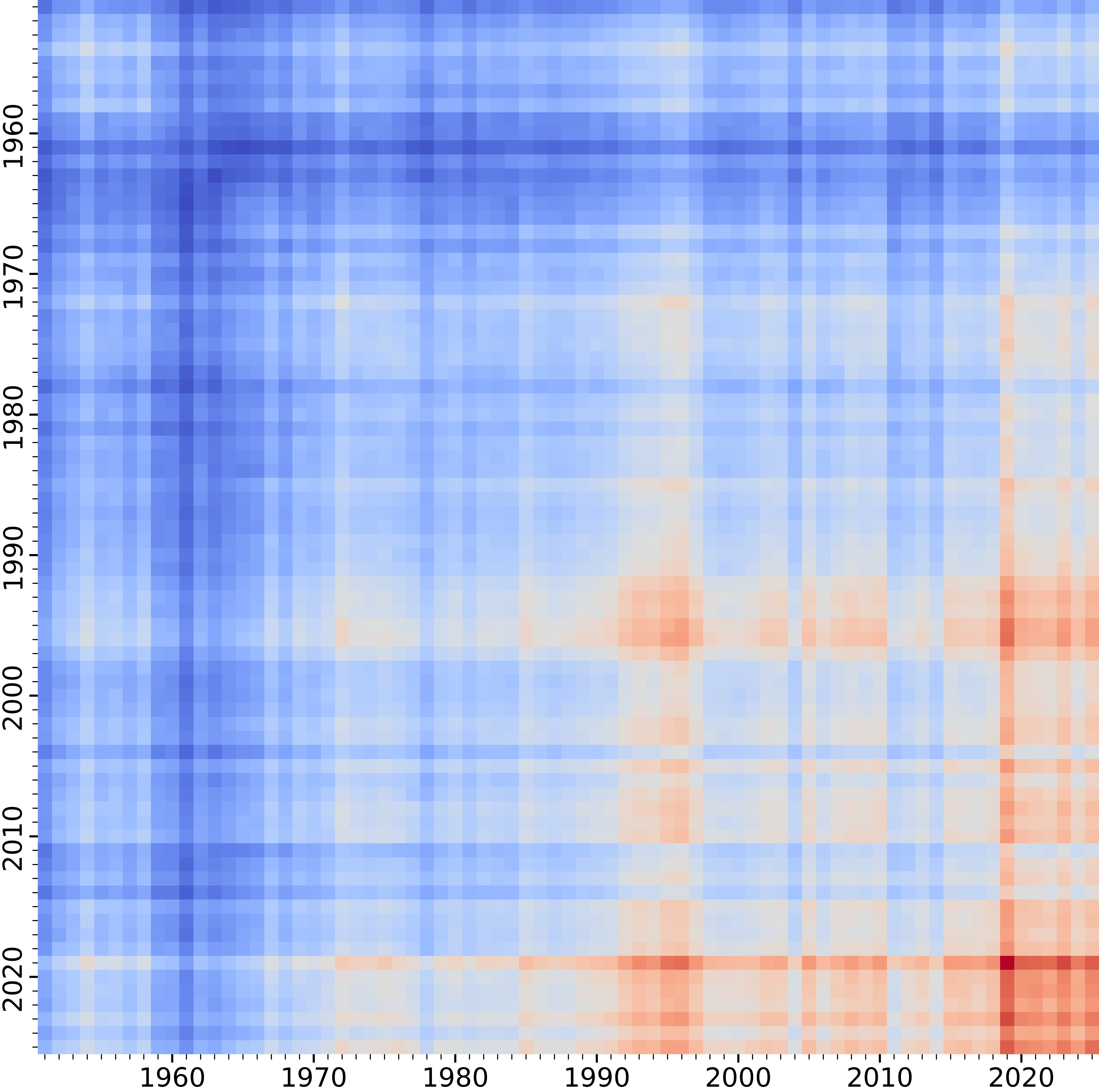}
    \caption{Matrix $\overline W^3_{mean}$, as defined in Section \ref{wordsbased},  shows the average overlap in significant normalized words (nouns, verbs, adjectives in base form) used in the lyrics between any two years. Warmer colors suggest greater lexical similarity.}
    \label{fig:word_normnorep_mean_norm}
\end{figure}

\section{Results and Discussion}\label{resultscom}

\begin{sidewaystable}
\caption{Examples of high lyrical similarity between segments. Original lyrics were in Italian and translated into English.}
\label{tab:lyric_similarity_segments}
\renewcommand{\arraystretch}{1.7}
\setlength{\tabcolsep}{5pt}
\vspace{0.3cm}
\begin{tabularx}{\textwidth}{|p{2.8cm}|p{2.9cm}|X|X|}
\hline
\textbf{Song 1} & \textbf{Song 2} & \textbf{Lyrics 1 (Translated)} & \textbf{Lyrics 2 (Translated)}\\
\hline

\makecell[l]{Gino Paoli \\ Un altro amore \\(2013)} &
\makecell[l]{G. Martinelli  \\Il gigante d’acciaio \\(2020)} &
\makecell[l]{~\\There won't be another love.\\ There won't be another time.\\ The one missing on my path\\ Was you, \\and no matter what\\ There won't be another love.\\~} &
\makecell[l]{There won't be\\ Another time, another time.\\ There won't be\\ Another time, yet again.} \\
\hline

\makecell[l]{Vasco Rossi \\ Vita spericolata \\(1983)} &
\makecell[l]{Tricarico \\ Vita tranquilla \\(2008)} &
\makecell[l]{I want a reckless life...\\ I want a rude life...\\ I want an excessive life...} &
\makecell[l]{~\\I... want a quiet life.\\ Because since I was born... \\it's been\\ Desperate... reckless...\\ Yet free... truly boundless.\\ I should... no, I shouldn't...\\~}  \\
\hline

\makecell[l]{E. De Crescenzo \\ Via con me \\(1983)} &
\makecell[l]{Anna Tatangelo \\ Libera \\(2015)} &
\makecell[l]{A cloud in the sky \\when the sun is out. \\ Swaying in the air, \\it drifts on its own.\\ Your fears drift away.\\ Adventures are forgotten.\\ But how can you stand \\being alone at night?} &
\makecell[l]{~\\Looking at fears.\\ Burning inside the sun.\\ And feeling that with you\\ I am free.\\ Free.\\ Like a cloud in the wind\\ That sways.\\ Unique, unique.\\ Like moonlight \\when it shines.\\~} \\
\hline
\end{tabularx}
\end{sidewaystable}

\begin{figure}[!h]
     \centering
     \includegraphics[width=0.6\linewidth]{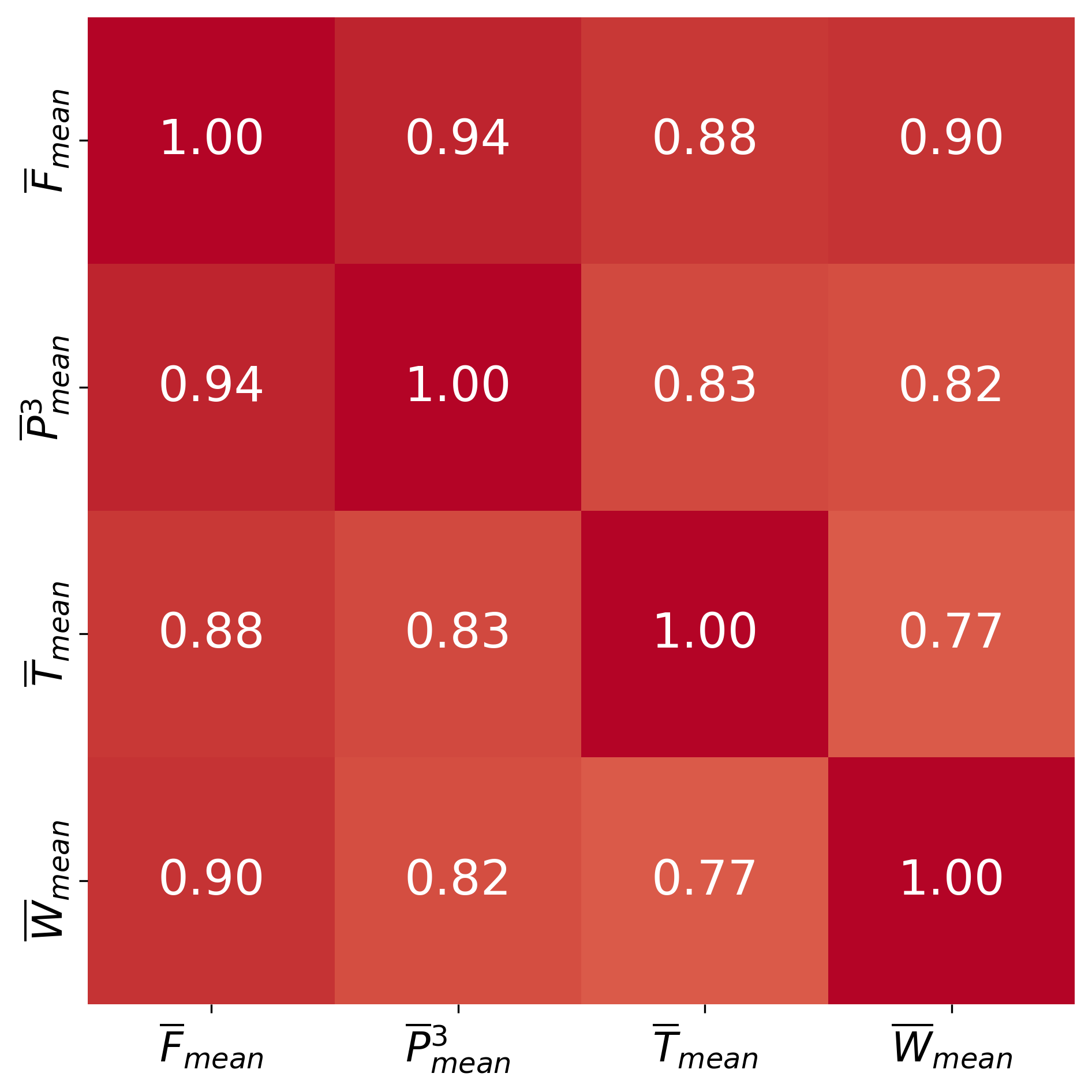}
    \caption{Pearson correlation coefficients between $\overline F_{mean}$, $\overline P^3_{mean}$, $\overline T_{mean}$ and $\overline W_{mean}$. The high values (warmer colors) suggest a strong positive agreement between the different lyrical similarity analysis methods (full text, portions, topics, and words), indicating consistent patterns in the results.}
     \label{fig:correlations}
 \end{figure}

To analyze the evolution of similarity (according to the different definitions) over years, we computed the following matrices $\overline F^{\alpha}_{\tau}, \overline P^{\alpha, 3}_{\tau}, \overline T^{\alpha}_{\tau}, \overline W^{\iota}_{\tau}$, for $\alpha \in \{i,o,c\}$, $\iota \in \{1,2,3,4\}$ and $\tau \in \{mean, med, 0.75\}$. Then, because the matrices $\overline F^{\alpha}_{\tau}$ (and analogously $\overline P^{\alpha, 3}_{\tau}, \overline T^{\alpha}_{\tau}, \overline W^{\iota}_{\tau}$ for $\iota \in \{1,2,3,4\}$),for  $\alpha\in E$, show the same pattern of similarities at the varying of $\tau \in \{mean, med, 0.75\}$, we decided to focus on the case $\tau = mean$. Moreover, to proceed with just one matrix for each type of similarity analysis considered, we defined the matrices $\overline W_{mean} := \sum_{\iota=1}^4 \overline W^{\iota}_{\tau}$, $\overline F_{mean}:= \sum_{\alpha \in E} \overline F^{\alpha}_{mean}$, $\overline P^3_{mean}:= \sum_{\alpha \in E} \overline P^{\alpha^3}_{mean}$ and $\overline T_{mean}:= \sum_{\alpha \in E} \overline T^{\alpha}_{mean}$. These latter matrices take into account all the different shades of similarity that the various embeddings or text normalization techniques capture in the different similarity analyses. To verify that all these different analyses lead to the same conclusions, we computed the Pearson correlation coefficients between the matrices defined above. The results, visible in Figure \ref{fig:correlations}, showing high values of correlations for all the possible pairs, suggesting that each analysis leads to the same conclusions. 

\begin{figure}[!htp]
    \centering
    \includegraphics[width=0.75\linewidth]{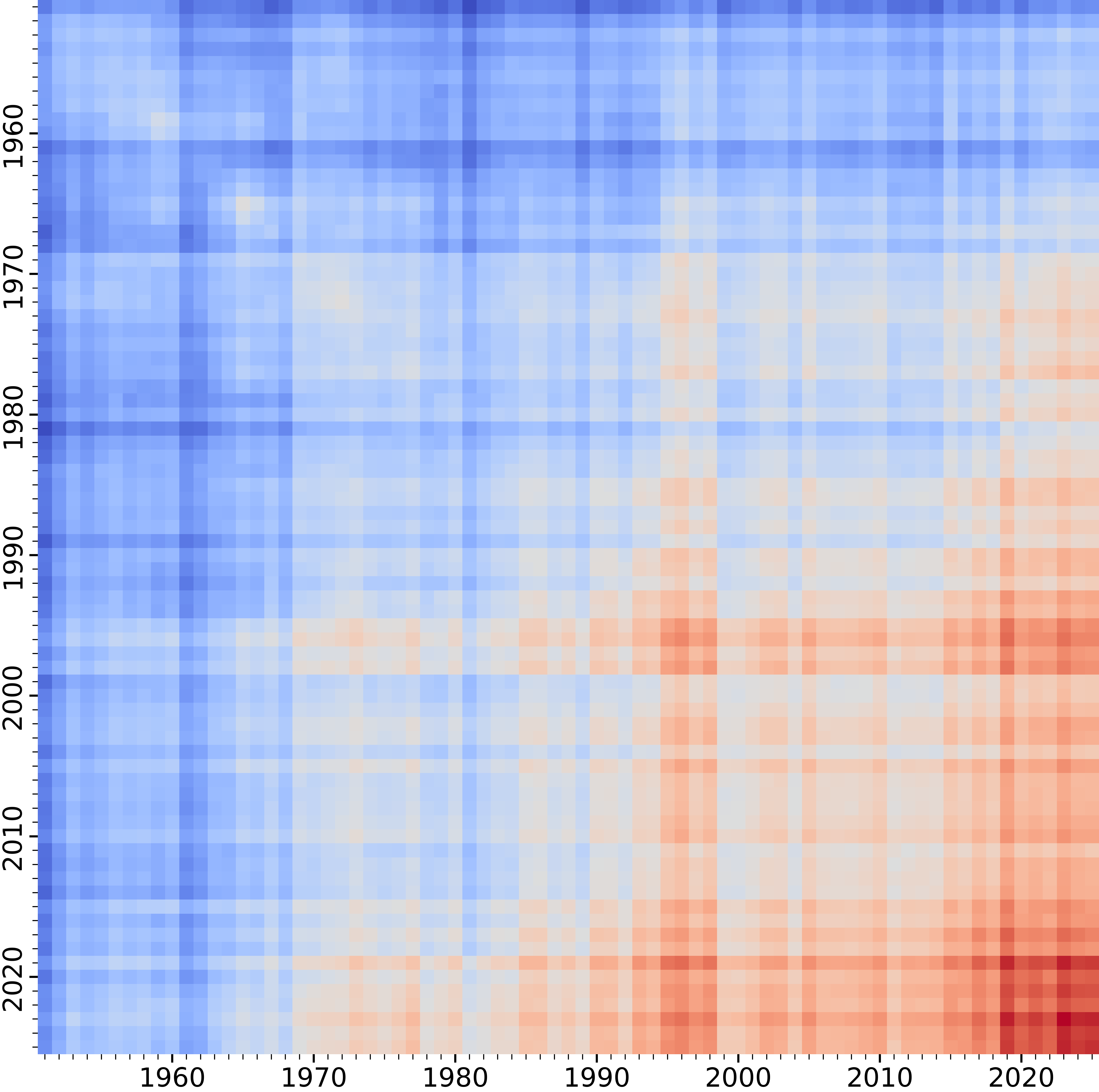}
    \caption{Matrix $\overline Z$, as defined in Section \ref{resultscom}, sum and normalize the results from all our analysis methods to show the general lyrical similarity between any two years, considering full text, key sections, themes, and word choice. Warmer colors indicate a stronger overall similarity.}
    \label{fig:summary}
\end{figure}

Hence, to visualize an overall similarity analysis across years, we defined the $|\mathcal{Y}| \times |\mathcal{Y}|$ matrix $Z = \overline F_{mean} + \overline P^3_{mean} + \overline T_{mean} + \overline W_{mean}$, which then we normalized in order that the minimum values is $0$ and the maximum $1$, obtaining the final matrix $\overline Z$, visible in Figure \ref{fig:summary}.

\begin{figure}[!htp]
    \centering
    \includegraphics[width=0.75\linewidth]{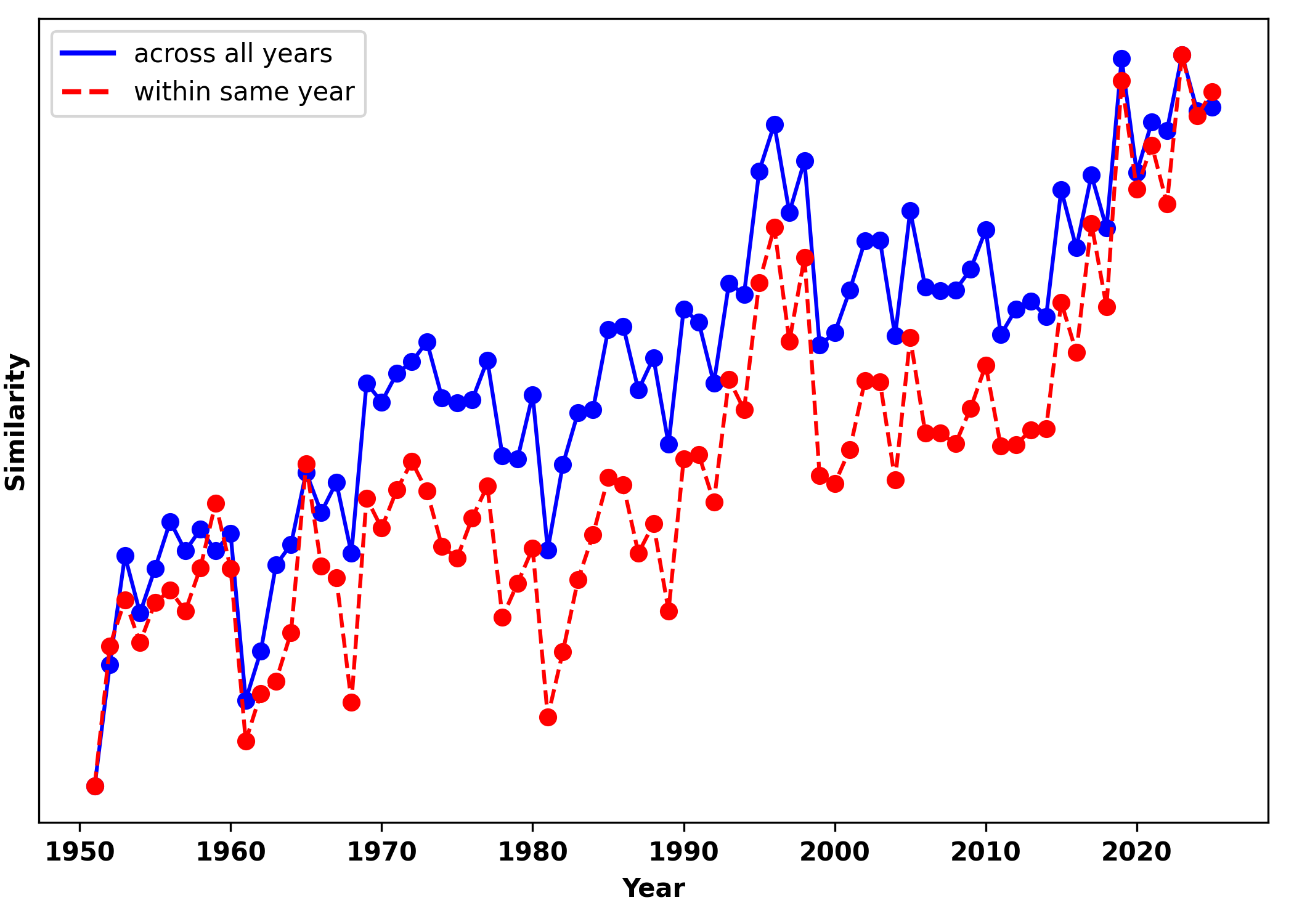}
    \caption{In dashed red the evolution across years of the average similarity within the single Sanremo Festival editions (diagonal of $\overline Z$). In blue the mean of the average similarities for each year against all the others (sum of $\overline Z$ rows).}
    \label{fig:similarity_combined}
\end{figure}

The matrix $\overline Z$ suggests that the correlation across the years is increasing with the years, becoming relatively bigger when computed on the last decades. Also the average similarity of the song presented in each year (diagonal of $\overline Z$) appears to be growing with the years, as it is clear from the plot in Figure \ref{fig:similarity_combined}, suggesting that the average similarity across the single editions of the festival is also growing with the years.

Moreover, it appears that the songs presented in the festival in the first years are less similar to all the others with respect to the recent ones. This is even more clear looking at the plots in Figure \ref{fig:similarity_combined}. 

\section{Conclusion and future work}\label{future}

In conclusion, this paper introduces a novel methodology, leveraging semantic embeddings and large language models, to analyze the evolution of lyrical diversity in musical collections. Applying this framework to the Sanremo Festival, a prominent Italian music competition, the study reveals a discernible trend towards increasing semantic homogeneity in lyrics over recent years. This finding is consistent across various similarity measures, including full-text, portion-based, topic-based, and word-based analyses, as confirmed by high Pearson correlation coefficients between these different approaches. This research not only provides a quantitative perspective on the evolution of Sanremo lyrics but also highlights the efficacy of semantic embeddings in uncovering nuanced trends within cultural artifacts.

This study opens avenues for future research:

\begin{itemize}
    \item \textbf{Developing a Python Framework}: We plan to release a Python framework based on our methodology that allows researchers to analyze how song similarity evolves based on a given variable (e.g., the release year of the songs). This tool would facilitate large-scale investigations into the temporal dynamics of lyrical similarity across different datasets.

    \item \textbf{Incorporating Musical Difference Metrics}: In \cite{kim2024computationalanalysislyricsimilarity}, the authors introduced the \textit{Musical Difference} metric, which has shown an even stronger correlation with human perceptions of lyrical similarity than standard semantic similarity measures. Future work should explore integrating this metric into our similarity analysis.

    \item \textbf{Validating the Methodology on External Datasets}: Since \cite{Parada-Cabaleiro2024} has made their dataset publicly available, it would be valuable to apply our methodology to their dataset and verify whether our approach yields similar findings. 
    
\end{itemize}

By addressing these directions, we aim to refine our methodology, improve its interpretability, and contribute to a broader understanding of how lyrical similarity evolves in different musical contexts.

\section*{Declarations}
\subsection*{Funding}
No funding was received for conducting this study. 
\subsection*{Competing interests}
The authors have no relevant financial or non-financial interests to disclose.
\subsection*{Conflict of interest}
On behalf of all authors, the corresponding author states that there is no conflict of interest.
\subsection*{Data availability}
The authors do not hold reproduction rights for the song lyrics; therefore, only the year of release, title, author, and interpreter of each song are provided (see Section \ref{dataemb}).

\bibliography{biblio}

\end{document}